# Visual Dominance and Emerging Multimodal Approaches in Distracted Driving Detection: A Review of Machine Learning Techniques

Anthony. Dontoh, *Member, IEEE*, Stephanie. Ivey, Logan. Sirbaugh, *Member, IEEE,* Andrews. Danyo, and Armstrong. Aboah, *Member, IEEE*

*Abstract* — Distracted driving continues to be a significant cause of road traffic injuries and fatalities worldwide, even with advancements in driver monitoring technologies. Recent developments in machine learning (ML) and deep learning (DL) have primarily focused on visual data to detect distraction, often neglecting the complex, multimodal nature of driver behavior. This systematic review assesses 74 peer-reviewed studies from 2019 to 2024 that utilize ML/DL techniques for distracted driving detection across visual, sensor-based, multimodal, and emerging modalities. The review highlights a significant prevalence of visual-only models, particularly convolutional neural networks (CNNs) and temporal architectures, which achieve high accuracy but show limited generalizability in real-world scenarios. Sensor-based and physiological models provide complementary strengths by capturing internal states and vehicle dynamics, while emerging techniques, such as auditory sensing and radio frequency (RF) methods, offer privacy-aware alternatives. Multimodal architecture consistently surpasses unimodal baselines, demonstrating enhanced robustness, context awareness, and scalability by integrating diverse data streams. These findings emphasize the need to move beyond visual-only approaches and adopt multimodal systems that combine visual, physiological, and vehicular cues while keeping in checking the need to balance computational requirements. Future research should focus on developing lightweight, deployable multimodal frameworks, incorporating personalized baselines, and establishing cross-modality benchmarks to ensure real-world reliability in advanced driver assistance systems (ADAS) and road safety interventions.

## I. INTRODUCTION

Road crashes are a critical global concern, with devastating human and economic consequences [1]. The World Health Organization (WHO) reports that approximately 1.19 million people lose their lives annually in road traffic accidents, with an additional 20 to 50 million sustaining non-fatal injuries, many of which lead to permanent disabilities [2]. Beyond the profound human toll, these accidents impose a significant economic burden on nations, accounting for an estimated 1 to 3 percent of Gross Domestic Product (GDP), with some countries facing costs as high as 6 percent [2]. These staggering impacts underscore the urgent need for continued efforts to enhance road safety and reduce the incidence of road crashes.

A crucial step in enhancing road safety is to understand the root causes of accidents. The National Highway Traffic Safety Administration (NHTSA) identifies distracted driving as one of the leading causes of road accidents [3], [4], [5]. For instance, the NHTSA reported in 2022 that distracted driving claimed over 3,000 lives and injured ~289,000 individuals in the United States alone [6]. Similarly, distracted driving accounts for 16% of fatal crashes in Australia and between 12% and 14% in Norway [7]. Beyond fatalities, distracted driving disrupts traffic efficiency, leading to erratic behaviors, traffic delays, increased fuel consumption, and exacerbated congestion [8], [9]. These aggravated impacts of distracted driving have caught the attention of several empirical studies [10].

Distracted driving encompasses any activity that diverts drivers' attention from driving, including cognitive, manual, visual, or auditory distractions, [6] highlighting the complexity of this behavior. This complexity has driven researchers to leverage both *simulation-based datasets* and *naturalistic driving study* (NDS) data to understand and mitigate its effects. In recent years, there has been a shift towards applying *machine learning* (ML) and *deep learning* (DL) techniques across these varied datasets for a comprehensive understanding of distracted driving behaviors [11], [12], [13], [14]. These advanced methods have demonstrated potential in capturing the intricate patterns of distracted driving behaviors in real-world scenarios, becoming a focal point for researchers [15].

Review papers by scholars have contributed significantly to our understanding of distracted driving behaviors, offering comprehensive overviews on various aspects of the subject. For instance, the study by Young et al. [16] focused *narrowly on in-vehicle distractions*, particularly cell phone use, while Papatheocharous et al. [15] examined *smartphone-based monitoring*, advocating for smartphone sensor-based data for distracted driving detection. Other scholarly review articles have focused on *specific aspects* such as *data collection* methods, *analysis techniques*, and *crash prevention strategies* [3], the *selection of metrics* for evaluating distracted driving countermeasures [17] and *the effectiveness of Convolutional Neural Networks* (CNNs) in real-time distraction detection [18]. Despite these valuable contributions, researchers have identified gaps in existing literature. Lansdown et al. [19], for instance, critiqued systematic review efforts, highlighting the *need for methodological rigor*, thus the need for more comprehensive analyses.

More recent review papers have addressed some of the methodological gaps highlighted by Lansdown [10]. Wang et al. [20] primarily focused on analyzing driver behavior using visual data obtained from in-vehicle cameras, reviewing techniques for monitoring driver states such as fatigue, drowsiness, and attentiveness using facial features, eye movements and head pose. Kashevnik et al. [21] in 2021 explored various methods, including the integration of visual, sensor, and physiological data to detect driver distraction, providing a framework for how multiple data sources can enhance detection accuracy and reliability. Hassan et al. [3] broadened the scope by discussing data collection methods and crash prevention strategies, highlighting the importance of naturalistic driving data in understanding real-world behaviors.

While these review studies provide valuable insights, they also exhibit certain limitations. They have often failed to explicitly acknowledge the prevailing emphasis on visual data in distracted driving detection research and the need to explore other approaches for a comprehensive distraction detection. For example, studies such as those by Wang et al. [20] concentrate on visual information, which highlights the field's primary focus. Similarly, broader reviews like those by Kashevnik et al. [21] and Hassan et al. [3] mention alternative modalities but allocate significantly greater attention to visual methods. This trend mirrors the current research landscape, where visual techniques surpass other approaches. This imbalance in existing reviews accurately represents the state of the field. However, it underscores the necessity for a review that explicitly addresses this visual predominance while exploring the emerging significance of other modalities and the need for greater attention to multimodal data approaches to distraction detection. Also, given the rapid evolution of techniques and approaches, there is a pressing need for an up-to-date analysis of current advances in distracted driving detection research.

This study addresses existing gaps by identifying and categorizing various machine learning (ML) and deep learning (DL) approaches for distracted driving detection. While previous reviews predominantly focus on visual detection methods, our review extends beyond this dominant modality to encompass sensory, multimodal, and emerging techniques but limited techniques. Our structure reflects the current research landscape, providing an up-to-date comprehensive overview of visual techniques and highlighting the significance of less prevalent approaches. The updated comparative analysis will emphasize the effectiveness of various methods in specific contexts and critically assess the limitations of single-modality systems despite their often-impressive accuracy metrics.

The study offers three key contributions to *methodological conceptualization* and *application* of ML and DL techniques in the context of distracted driving. First, it shows the impact of visual dominance on the current research landscape while highlighting promising developments in alternative modalities. Second, it provides clear guidance on the limitations of single-modal analysis despite the high reported accuracies, particularly in context-sensitive scenarios where multimodal approaches may present future advantages. Third, it synthesizes recent studies that have made significant strides, delivering a comprehensive overview of distracted driving research's current state and direction. Ultimately, this review aims to enhance road safety by advancing detection methods for distracted driving in the next generation of Advanced Driver Assistance Systems (ADAS) and informing more effective mitigation strategies.

This systematic review concentrates on peer-reviewed studies that employed machine learning (ML) and deep learning (DL) techniques to address the issue of distracted driving. The research thoroughly searched multiple databases, including IEEE Xplore, Google Scholar, TRB, Scopus, and PubMed. These platforms encompass various relevant fields, such as electrical engineering, computer science, traffic safety, and health sciences. This comprehensive approach ensured a multidisciplinary perspective on the technologies utilized to detect and mitigate distracted driving, highlighting the latest advancements in computational methods and their practical applications to enhance road safety.

The remainder of this paper is organized as follows: Section II presents the methodology employed for identifying, selecting, and analyzing relevant studies. Section III categorizes the literature reviewed based on the primary data modalities utilized—namely, visual, sensor-based, multimodal, and emerging approaches. Within each modality, representative ML and DL techniques are discussed, followed by a summary and comparative analysis highlighting key strengths and limitations. Section IV concludes the paper with a synthesis of key findings and outlines future research directions.

## II. METHODOLOGY

This review follows established systematic review protocols [22], [23], [24], [25], [26], [27] coupled with expert consultation in computer vision and driver behavior [28]. The methodology comprises key steps, including study selection, inclusion and exclusion criteria, and final screening to ensure relevance of reviewed papers.

### A. Study Selection

A comprehensive literature search was conducted across multiple academic databases, including IEEE Xplore, Google Scholar, TRB, Scopus, and PubMed. The search was restricted to peer-reviewed journal articles published between January 1, 2019, and December 31, 2024, to capture recent advancements in ML and DL techniques applied to distracted driving detection. Seminal studies published prior to this period were selectively included if they made significant conceptual contributions to the development of the field.

These databases were selected to ensure broad and relevant coverage of the literature across disciplines central to this review. Scopus offers a comprehensive and curated index of scientific journals and conference proceedings, particularly valuable for multidisciplinary research. Google Scholar provides broad access to academic materials across various formats and domains, including engineering and telehealth. PubMed was included for its high-quality indexing of biomedical and health-related literature, which complements the technical focus of platforms like Scopus and IEEE Xplore. Together, these databases minimize selection bias and enhance the feasibility and comprehensiveness of the review.

The search strategy incorporated a combination of keywords such as "distracted driving detection," "machine learning," "deep learning," "multimodal data," "naturalistic driving data," and "simulation-based studies." Boolean operators AND/ OR were applied to broaden the search scope and retrieve relevant literature aligned with the objectives of this review.

*B. Inclusion and Exclusion Criteria*

The study ensured that only papers relevant to the study's objectives were included by introducing the following inclusion and exclusion criteria:

*Inclusion Criteria:* The review included studies that applied ML or DL techniques to distracted driving detection. Both public and private datasets were considered to provide comprehensive coverage. Studies incorporating distinct data modalities, such as visual, auditory, sensory, or multimodal approaches, were prioritized. Novel modalities offering unique perspectives on distraction detection were also considered.

*Exclusion Criteria:* Studies focused solely on crash prevention without applying machine learning or deep learning were excluded. Articles not available in full text or not published in English were also omitted. Additionally, dissertations, grey literature, non-peer-reviewed sources, and unpublished studies were not considered.

*C. Screening Process*

The screening process followed a structured, multi-phase review in accordance with the Preferred Reporting Items for Systematic Reviews and Meta-Analyses (PRISMA) guidelines. The initial search yielded 606 records from aforementioned databases with one additional article identified through other sources, bringing the total to 607 records.

All identified records were imported into Mendeley Reference Manager, where 2 duplicates were automatically detected and removed. The remaining 605 records were subjected to title and abstract screening, during which 285 irrelevant records were excluded based on the study's focus and relevance to machine learning (ML) or deep learning (DL) applications for distracted driving detection.

A total of 321 articles remained for screening. Of these, 137 full-text articles could not be retrieved, often due to access limitations or broken links. This left 100 full-text articles to be assessed for eligibility.

The full-text screening was conducted by all authors of the paper. Each group independently reviewed all 100 eligible articles using a standardized inclusion checklist based on the review's objectives. Discrepancies between groups were resolved through discussion to ensure consistency and alignment with the inclusion criteria.

Of the 100 full-text articles reviewed, 26 were excluded for reasons such as lack of ML/DL focus, missing evaluation metrics, or non-English language. Ultimately, 74 studies were included in the final review, along with 1 additional report, as reflected in the PRISMA flow diagram as shown in Figure 1, developed in accordance with the guidelines by Haddaway et al.[29].

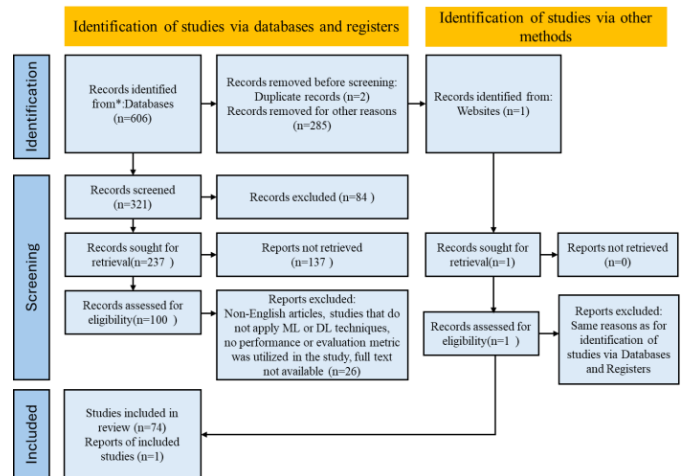

**Fig. 1.** PRISMA Flow Diagram: Study Identification and Screening Process.

### III. RESULTS OF THIS STUDY AND DISCUSSION

Research on distracted driving detection has advanced considerably with the proliferation of sensing technologies and computational models. This section categorizes detection strategies by primary data modality, including visual, sensor-based, multimodal, and emerging approaches—and evaluates their methodological trends, performance characteristics, and practical limitations.

Visual methods dominate the literature, often achieving high accuracy using advanced deep learning and computer vision techniques [30], [31], [32]. However, their reliance on appearance-based cues limits robustness under poor lighting, occlusions, or in detecting cognitive distraction [33], [34], [35].

Sensor-based approaches, including vehicle telemetry and physiological monitoring, offer complementary insights by capturing temporal dynamics and internal states [14], [36], [37]. Nevertheless, these models often suffer from reduced generalizability due to reliance on simulated data, constrained sensor availability, or limited real-world validation.

The review highlights trade-offs between detection accuracy, computational efficiency, and deployment feasibility across modalities. Notably, the disparity in methodological sophistication, particularly between visual and non-visual modalities, suggests untapped opportunities for cross-modality innovation [11], [38].

To address the multifaceted nature of distraction, this section ultimately advocates for multimodal approaches that fuse visual, behavioral, and physiological signals. These systems show superior context awareness and generalization across diverse driving scenarios [13], [14], [39]. The following subsections detail representative methods within each modality, assessing their strengths, limitations, and potential for real-world implementation.

*A. Visual Data Approaches*

Visual data from in-vehicle cameras has emerged as the *most widely* used modality for distracted driving detection [3], capturing real-time cues such as *facial expressions, eye gaze, head posture, arm and body movements* [40]. These indicators enable detection of various distractions, including manual (e.g., phone use) and visual (e.g., eyes off the road) behaviors. The rich information captured through visual data has led researchers to explore diverse ML and DL approaches for effective analysis. Common techniques include CNNs for image processing, temporal models like *Long Short-Term Memory* (LSTM) networks for sequential analysis, and *hybrid architectures* combining multiple approaches for improved performance. These methods have demonstrated capability in processing complex visual information to detect and classify different types of distracted behaviors in both offline analysis and real-time applications. This section examines how these techniques have been applied to *visual data* for distracted driving detection, highlighting key methodological approaches and implementations. Figure 2 gives an overview of the various visual data approaches captured in this review.

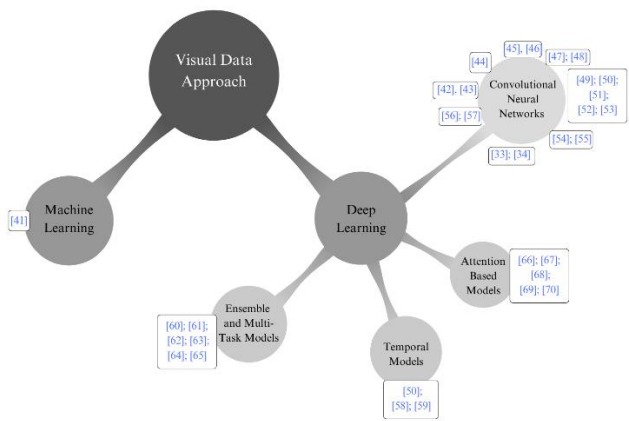

**Fig. 2.** Summary of Visual Data Approaches

*ML and DL-Based Approaches*

This section outlines the primary strategies applied to visual data, categorized into traditional machine learning and deep learning techniques.

**Machine Learning (ML) Approaches**

Although deep learning methods dominate recent efforts as depicted in Figure 2, traditional ML approaches have also been explored for visual distracted driving detection. Gong and Shen [41] proposed a model that combines pose-based features with an Adjustable Distance-Weighted K-Nearest Neighbors (ADW-KNN) classifier. Using OpenPose for skeletal keypoint extraction, the model achieved 94.04% accuracy and 50 FPS on the State Farm Distracted Driver Detection (SFDDD) dataset, demonstrating promise for real-time deployment.

However, the method's reliance on iterative parameter tuning for $k$ and $p$ values increases computational overhead, and the need for larger $k$ values than standard KNN may limit scalability on larger datasets. These challenges highlight the need for more adaptable and scalable ML frameworks in future work.

**Deep Learning Approaches**
**Convolutional Neural Networks (CNNs)**

CNNs remain foundational in visual distracted driving detection due to their capacity to extract complex spatial features from in-cabin camera data [42], [43]. Over the past five years, CNN-based models have evolved in two primary directions: *maximizing accuracy* through region-focused and transfer learning strategies and *minimizing computational complexity* for real-time or edge deployment.

Region-specific CNN architectures have proven particularly effective at localizing distraction-related behaviors. Wang et al. [44] applied Faster R-CNN to focus on the driver's upper body and steering wheel, boosting accuracy (96.97%) by filtering irrelevant background features. Zheng et al. [45] introduced CornerNet-Saccade, achieving 97.2% accuracy by isolating face and hand regions. These methods, however, required significant computational resources and struggled under uncontrolled conditions, such as poor lighting or occlusion.

More efficient CNN variants have emerged to improve both scalability and performance. Sajid et al. [46] employed EfficientDet-D3, which integrates hierarchical scaling and Bidirectional Feature Pyramid Network (BiFPN) feature fusion, to achieve a mAP of 99.16%, outperforming Faster R-CNN and You Only Look Once (YOLOv3) in classifying manual distractions. Similarly, Peruski et al. [47] demonstrated that EfficientNetB7 could reach an AUC of 1.00 when trained on high-quality datasets. While these results are impressive, their dependence on clean, well-annotated input limits transferability to varied environments.

Efforts to reduce model complexity without sacrificing accuracy have led to the development of lightweight CNNs. Sahoo et al. [48] presented a SqueezeNet-based model that achieved 99.93% accuracy on the SFDDD dataset and was optimized for resource-limited devices such as Raspberry Pi 4B. Qin et al. [49] developed the D-HCNN model, which used Histogram of Oriented Gradients (HOG) based preprocessing to suppress background noise and achieved up to 99.87% accuracy. However, the complexity of convolutional layers and limited adaptability to real-world scenarios constrained scalability.

Liu et al. [50] introduced a knowledge-distilled lightweight CNN with only 0.42 million parameters, highlighting the feasibility of compression while maintaining high accuracy (99.86%). Although they proposed expansion into a 3D CNN for temporal analysis, such extensions remain conceptual.

López and Arias-Aguilar [51] adopted a TensorFlow Lite-based model optimized for embedded devices, achieving 91% accuracy. However, their evaluation under normal weather conditions left questions about generalizability in adverse environments.

Transfer learning has also proven pivotal in leveraging pre-trained visual models. Anand et al. [52] and Supraja et al. [53] employed VGG16 to achieve over 90% accuracy, integrating real-time alert systems into their frameworks. Zhang and Ke [33] used regularization and augmentation techniques with VGG16 and ResNet50 to achieve 98.5% accuracy while mitigating overfitting. Oliveira and Farias [54] compared VGG19, DenseNet161, and InceptionV3, concluding that end-to-end optimization, particularly with DenseNet161, yielded superior results (88.83% accuracy). Their study also revealed that feature extraction-only strategies significantly underperformed, dropping test accuracy to 57.41% and below 30% in some cases. Nur et al. [55] further demonstrated the strength of transfer learning with a modified VGG16 model, achieving 99.86% accuracy using visual cues such as head orientation, hand position, and body posture. While the results were promising, the study's limited dataset size and lack of integration into a real-world monitoring system highlighted persistent gaps in applicability.

Goel et al. [56] demonstrated practical applications of transfer learning by deploying MobileNetV2 within a web-based interface, achieving 93.8% accuracy and showcasing feasibility in user-facing applications. However, like many others, their model lacked coverage of cognitive distractions. Vaegae et al. [34] compared VGG16 and ResNet50, highlighting edge deployment challenges and advocating for multimodal inputs to enhance robustness.

Finally, fusion-based approaches integrating CNNs with detection pipelines have shown strong potential. Fan and Shangbing [57] developed a multi-scale fusion network that combined global features from ResNet-50 with localized cues extracted via YOLOv5. Their model achieved 95.84% accuracy on the AUC dataset and demonstrated improved differentiation of visually similar behaviors, such as texting versus calling. Despite these advantages, the system's computational complexity introduced challenges for real-time deployment.

Across these efforts, CNN-based models have achieved state-of-the-art performance on benchmark datasets and are well suited for detecting overt, physical distractions. Reported accuracy across reviewed CNN studies ranges from 91% in lightweight, embedded applications [51] to over 99.9% in optimized architectures such as SqueezeNet [48] and EfficientNetB7 [47], with most models consistently achieving above 95% accuracy under controlled conditions. However, *their overreliance on visual appearance and single-frame interpretation* limits their capacity to detect nuanced or cognitive distractions. Moreover, the lack of diverse training data, especially regarding demographics, driving conditions, and camera perspectives, impedes generalizability. These issues collectively point toward the need for systems to integrate additional contextual signals beyond visual data, a challenge that more holistic, multimodal models are better positioned to address.

Figure 3 provides the modular pipeline for CNN-based distracted driving detection approaches while Table 1 provides a summary of performance.

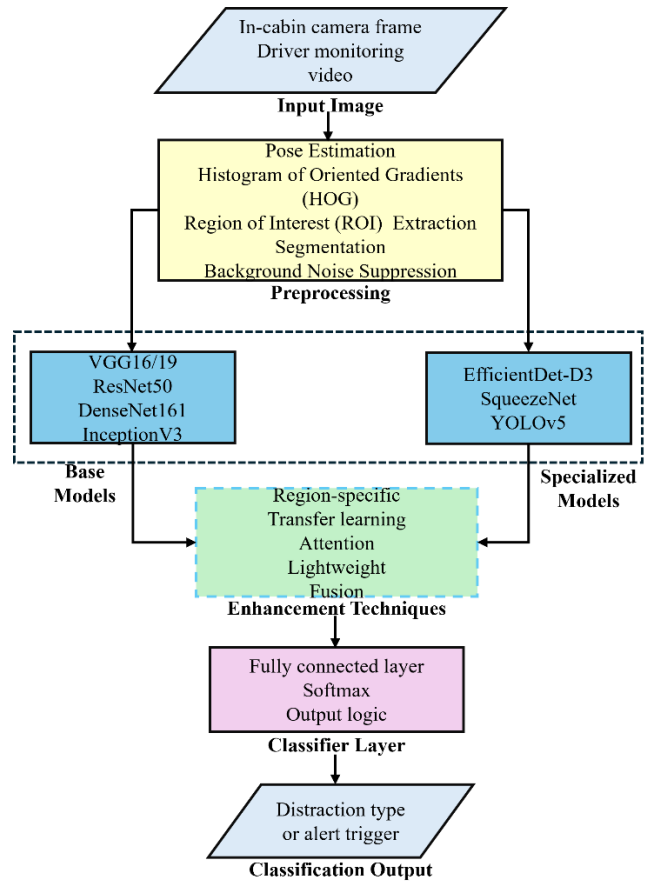

**Fig. 3.** Modular Pipeline for CNN-based distracted driving detection.

TABLE I
SUMMARY OF CNN-BASED APPROACHES

| Study | Architecture | Performance | Key Notes |
|---|---|---|---|
| Wang et al. [44] | Faster R-CNN | 96.97% (AIAC) | Region-specific filtering of facial features |
| Zheng et al. [45] | CornerNet-Saccade | 97.27% (Private) | Isolated facial hands for higher precision |
| Sajid et al. [46] | EfficientDet-D3 | 99.16% (Private) | Outperformed YOLOv3; robust for manual tasks |
| Peruski et al. [47] | EfficientNetB7 | 1.00 AUC (Private) | Input quality boosts result |
| Sahoo et al. [48] | SqueezeNet | 99.93% (SFDDD) | Real-time friendly; edge-deployable |
| Qin et al. [49] | D-HCNN | 99.87% (Private) | HOG-based preprocessing; noise suppression |
| Liu et al. [50] | Knowledge-distilled Lightweight CNN | 99.86% (Private) | Model compression; minimal parameters |
| López & Arias-Aguilar [51] | TensorFlow Lite CNN | 91.00% (Private) | Streamlined for embedded deployment |
| Anand et al. [52] | VGG16 | >90.00% (Private) | Real-time alert system |
| Supraja et al. [53] | VGG16 | >90.00% (Private) | Distraction-aware attention systems |
| Zhang & Ke [33] | VGG16, ResNet50 | 98.50% (Private) | Data augmentation and regularization |
| Oliveira & Farias [54] | VGG19, DenseNet161, InceptionV3 | 88.83% (Private) | DenseNet161 best; feature-only strategies underperformed |
| Nur et al. [55] | VGG16 | 99.86% (Private) | Posture, hand, and head-based attention |
| Goel et al. [56] | MobileNetV2 | 93.80% (Private) | Web-based deployment |
| Vaegae et al. [34] | VGG16, ResNet50 | 93.60% (Private) | Highlights edge deployment barriers |
| Fan & Shangbing [57] | ResNet-50 + YOLOv5 | 95.84% (AUC) | Multi-scale fusion of localized regions |

**Temporal Models**

Temporal models have emerged as a critical advancement in distracted driving detection, addressing the limitations of frame-level CNN classifiers by capturing dynamic patterns across time. While CNNs excel at identifying spatial features from still images, they often fail to detect context-driven or cognitive distractions that unfold over sequences of frames. Temporal architectures such as ConvLSTMs, hybrid CNN-RNNs, and Transformer-based models offer a robust solution to this problem by modeling driver behavior as a time-evolving process.

One of the earliest efforts to introduce temporal depth was presented by Liu et al. [50], who extended their lightweight knowledge-distilled CNN into a *3D CNN* for spatial-temporal driver behavior recognition. Using the Drive&Act dataset, the model achieved notable performance improvements with only 2.03 million parameters, striking a balance between efficiency and accuracy. However, the architecture required careful kernel tuning and remained computationally demanding during training, limiting its broader applicability for real-time vehicle deployment.

Hybrid CNN-RNN models have been proposed to capture long-term dependencies more effectively. Kumar et al. [58] designed a genetically optimized ensemble comprising six deep models, including a CNN-BiLSTM hybrid that paired InceptionV3 with a bidirectional LSTM. This ensemble achieved *96.37% accuracy on the AUC dataset* and an exceptional *99.75% on SFDDD*, demonstrating its capability to detect subtle distractions like reaching behind or interacting with a mobile device. Despite its success, the ensemble's complexity raises concerns about real-time feasibility and requires evaluation of embedded systems for practical use. Furthermore, the class overlap between visually similar activities (e.g., grooming vs. talking) remained a source of misclassification.

The latest generation of temporal models has adopted *Transformer-based architectures* for their superior capacity to model long-range dependencies without the limitations of recurrent connections. Shi [59] introduced the DKT framework, which integrates DWPose keypoint detection with a multi-transformer module and a High- and Low-Frequency Multi-Transformer Attention (HLMSA) mechanism. The model achieved *98.03% accuracy on the 100-Driver Dataset* and *73.88% cross-dataset generalization on SFD2*, outperforming several state-of-the-art alternatives. Key innovations included jitter suppression through Kalman filtering and frequency-aware attention. However, the HLMSA module contributed more to computational efficiency than accuracy gains, and the model's reliance on pre-extracted key points posed generalization challenges in noisy or unstructured environments.

Across these efforts, temporal models have effectively addressed limitations found in CNN-only approaches. They are robust in handling distractions involving prolonged gaze shifts, complex arm movements, or sequences of subtle actions. Reported performance among temporal models ranged from *96% to nearly 99.8%* under controlled conditions, with generalization drops when tested across datasets or under varied lighting and occlusion conditions. While most studies adopted temporal modeling as an extension of CNN architectures, the increasing use of attention mechanisms and lightweight temporal units reflects a growing emphasis on computational efficiency and real-time readiness.

Despite their strengths, temporal models still struggle with deployment challenges, including the need for annotated video sequences, increased inference time, and limited availability of large-scale temporal datasets. Furthermore, most implementations focus exclusively on visual cues, leaving cognitive and environmental contexts unmodeled. These shortcomings underscore the need for multimodal approaches integrating temporal visual analysis with sensor data, physiological cues, or auditory signals to detect distracted driving.

Figure 4 provides the generalized architecture for temporal modeling in distracted driving detection while Table 2 summarizes performance across reviewed studies.

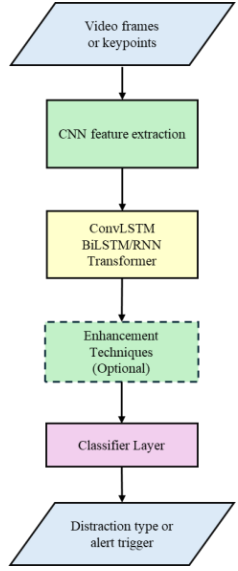

**Fig. 4.** Generalized Architecture for Temporal Modeling in Distracted Driving Detection.

TABLE 2.
SUMMARY OF TEMPORAL MODELING APPROACHES

| Study | Architecture / Model Type | Performance | Key Notes |
|---|---|---|---|
| Liu et al. [50] | 3D-CNN (Knowledge-distilled) | Not specified | Spatial-temporal extension of lightweight CNN; 2.03M parameters; kernel tuning needed |
| Kumar et al. [58] | CNN-BiLSTM (InceptionV3 + BiLSTM) | 96.37% (AUC); 99.75% (State Farm) | Genetically optimized ensemble; sensitive to subtle distractions; high computational cost |
| Shi [59] | Transformer-based (DKT framework with HLMSA) | 98.03% (100-Driver); 73.88% (SFD2 generalization) | Kalman filtering and frequency-aware attention; relies on keypoint pre-processing |

**Ensemble and Multi-task Models**

Ensemble and multi-task learning strategies have emerged as powerful tools in distracted driving detection, addressing limitations in single-model architectures by fusing the outputs or learned representations from multiple models. These methods improve robustness, mitigate biases associated with specific CNN backbones, and enhance generalizability across diverse driver behaviors.

Hybrid CNN ensembles remain the most used form, combining different convolutional networks to maximize the detection of subtle distractions. Huang et al. [60] introduced a cooperative transfer learning approach integrating ResNet50, InceptionV3, and Xception, achieving 96.74% accuracy on the State Farm dataset. This ensemble captured subtle hand and body gestures well but struggled in low-light conditions and with right-sided camera placement bias.

Similarly, Subbulakshmi et al. [61] built a broader ensemble incorporating ResNet50, VGG16, and DenseNet121. Their model achieved 98.92% accuracy, excelling at distinguishing between overlapping behavior classes. However, the study acknowledged the need for advanced ensemble fusion techniques and improved augmentation strategies to enhance generalization under varied driving contexts.

Weighted averaging ensembles provide an alternative strategy for integrating model outputs. Mollah et al. [62] proposed a SoftMax-weighted ensemble of DenseNet121 and MobileNet, which achieved 99.81% accuracy while maintaining only 10.2 million parameters. This compact architecture demonstrated efficient feature diversity but required improved regularization to combat overfitting.

Stacking-based ensembles, such as the E2DR model developed by Aljasim and Kashef [63], used complementary strengths of VGG16 and ResNet50, achieving 92% accuracy while maintaining low inference latency. Despite its modest accuracy, E2DR was optimized for real-time scenarios. However, ensemble construction from scratch proved computationally demanding, and the model lacked integration with surrounding vehicle or road context.

Draz et al. [64] introduced an innovative object-pose ensemble, combining Faster R-CNN for object detection with pose estimation via Intersection over Union (IoU) logic. The model achieved 92.2% accuracy and showed promise in localizing distractions involving handheld objects. Still, its focus on visual cues alone and sensitivity to environmental factors limited scalability, echoing challenges shared by other visual-only ensemble systems.

Multi-task learning frameworks represent a growing area of innovation. Liu et al. [65] proposed the Triple-Wise Multi-Task Learning (TML) model, which incorporated triplet-based image sampling to improve both classification and feature discrimination. TML achieved 96.3% accuracy on the AUC dataset and 66.9% cross-dataset performance on Drive&Act, showcasing enhanced generalization. However, the model required approximately 90 million parameters, limiting its suitability for embedded deployment. Its reliance on input-positive sample averaging to achieve optimal performance also highlighted a potential inefficiency in training dynamics.

Across these approaches, ensemble and multi-task systems demonstrate distinct advantages in accuracy, robustness, and generalization. Yet, their computational demands, overfitting risks, and narrow modality scope remain critical bottlenecks. Most models continue to rely solely on visual features, missing opportunities to detect cognitive or auditory distractions. Integrating these architectures into multimodal frameworks could further elevate their utility in real-world applications.

Figure 5 provides the generalized architecture for ensemble and multitask approaches in distracted driving detection while Table 3 summarizes performance across reviewed studies.

Broadly, attention mechanisms are implemented either by integrating them into existing object detection frameworks (for example, You Only Look Once or YOLO) or embedding channel, spatial, or transformer-based attention modules within convolutional neural network (CNN) architectures.

YOLO-based attention enhanced models are especially notable for their balance of speed and accuracy. Du et al. [66] augmented YOLOv8n with Ghost Convolution (GhostConv) layers, Bidirectional Feature Pyramid Networks (BiFPN), and Simple Attention Module (SimAM), reducing computational demands by 36.7 percent while covering 14 distraction categories including night time driving. Li et al. [67] developed the Attention Based Driver Behavior Learning Model (AB DLM), embedding Squeeze and Excitation (SE) attention and BiFPN into YOLOv5s, achieving 95.6 percent mean average precision (mAP) at 71 frames per second (FPS), outperforming lightweight variants such as YOLOv4 Tiny and PP YOLO Tiny. Similarly, Du et al. [68] applied Convolutional Block Attention Module (CBAM) and BiFPN to YOLOv5, attaining 91.6 percent precision and 89.2 percent mAP across 14 distraction types. These approaches demonstrated strong localization performance and real time viability, though they remained sensitive to lighting variations and occlusions.

In contrast, other models integrate attention directly within convolutional pipelines. Ai et al. [69] proposed the Dual Attention Convolutional Neural Network (DACNN), which combined spatial and channel attention to improve subtle behavior discrimination, achieving 95.4 percent accuracy. Li et al. [70] introduced the Spatial Weighted Attention Module (SWAM), a self-attention mechanism tailored for visual distraction detection. Tested on the American University in Cairo Version 2 (AUC V2) dataset, it achieved 93.97 percent accuracy while offering long range dependency modeling and enhanced focus on key visual regions. Both methods highlight the interpretability benefits of attention, though the added architectural complexity poses deployment challenges on edge devices.

Overall, attention-based models report high performance, with accuracy ranging from *91.6 percent to 95.6 percent,* and contribute to model interpretability and robustness under varied visual conditions. Nonetheless, their reliance on controlled datasets, elevated computational complexity, and limited cross dataset evaluation hinder generalizability. Furthermore, most implementations focus exclusively on visual cues, missing broader contextual signals such as auditory, physiological, or vehicular data. Future directions should explore the integration of attention mechanisms within multimodal architectures to capture a fuller spectrum of driver behavior and distraction sources.

Figure 6 provides the generalized architecture for attention-based models in distracted driving detection while Table 4 summarizes performance across reviewed studies.

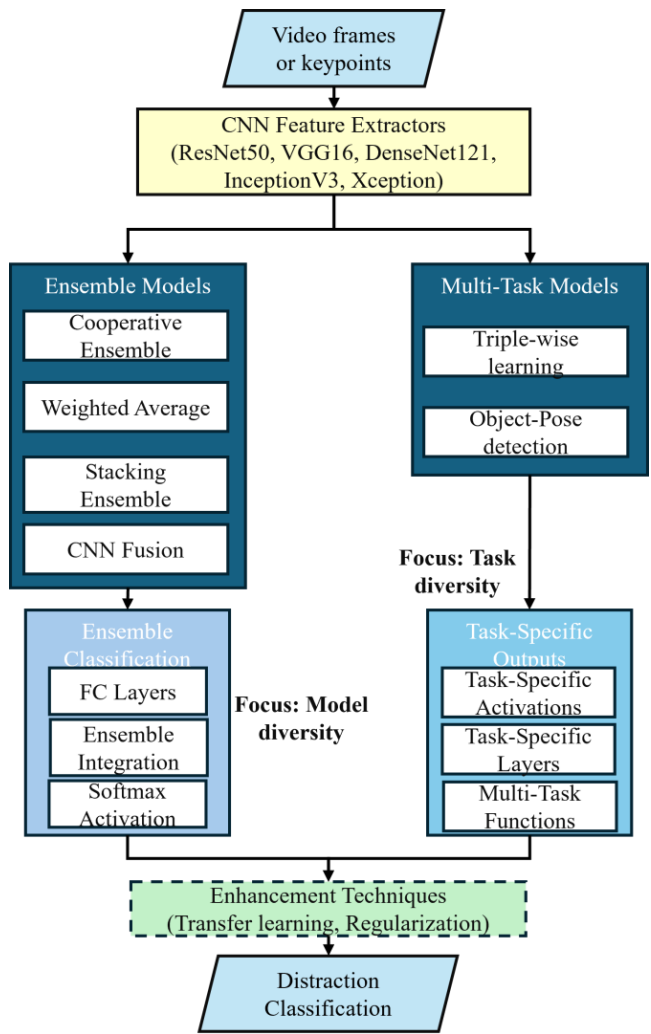

**Fig. 5.** Conceptual architecture depicting ensemble and multi-task learning strategies for distracted driving detection.

TABLE 3.
SUMMARY OF ENSEMBLE AND MULTI-TASK APPROACHES

| Study | Architecture / Strategy | Performance | Key Notes |
|---|---|---|---|
| Huang et al. [60] | Cooperative Transfer Learning (ResNet50, InceptionV3, Xception) | 96.74% (State Farm) | Captures subtle body cues; low-light and camera placement bias limitations |
| Subbulakshmi et al. [61] | Ensemble (ResNet50, VGG16, DenseNet121) | 98.92% (Private) | Strong class separation; needs better fusion and augmentation strategies |
| Mollah et al. [62] | SoftMax-Weighted Ensemble (DenseNet121, MobileNet) | 99.81% (Private) | Compact with 10.2M params; overfitting risk due to limited regularization |
| Aljasim & Kashef [63] | Stacked Ensemble (VGG16 + ResNet50, E2DR) | 92.00% (Private) | Low latency; no road context integration; high training cost |
| Draz et al. [64] | Object-Pose Fusion (Faster R-CNN + IoU pose) | 92.20% (Private) | Handheld object localization; visual-only limits generalization |
| Liu et al. [65] | Triple-Wise Multi-Task Learning (TML) | 96.30% (AUC); 66.90% (Drive&Act) | Triplet sampling improves generalization; large model size (~90M params) |

**Attention Based Models**

Attention mechanisms have emerged as a crucial enhancement in visual distracted driving detection, enabling models to prioritize semantically relevant regions such as the driver's face, hands, and posture while suppressing background noise. These methods improve both classification accuracy and computational efficiency, particularly in real time applications.

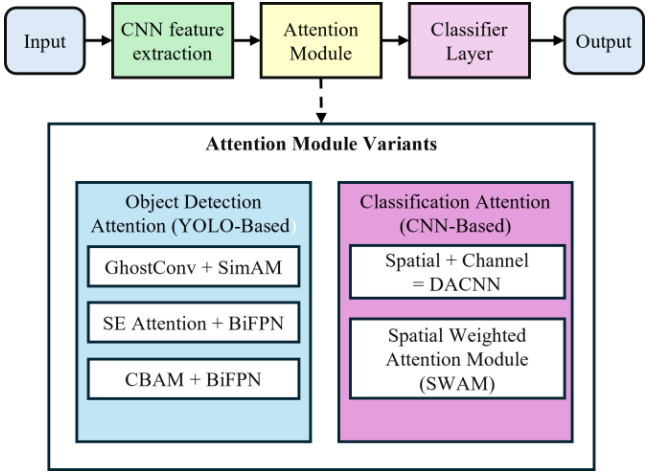

**Fig. 6.** Modular Architecture of Attention-Enhanced CNNs for Distracted Driving Detection.

TABLE 4.
SUMMARY OF ATTENTION BASED MODEL APPROACHES

| Study | Architecture / Attention Strategy | Performance | Key Notes |
|---|---|---|---|
| Du et al. [66] | YOLOv8n + GhostConv + BiFPN + SimAM | 91.6% precision; 14 distraction types (Private) | Reduced computation by 36.7%; strong real-time localization |
| Li et al. [67] | AB DLM (YOLOv5s + SE Attention + BiFPN) | 95.6% mAP @ 71 FPS (Private) | Outperformed YOLOv4 Tiny and PP YOLO Tiny; fast and accurate |
| Du et al. [68] | YOLOv5 + CBAM + BiFPN | 91.6% precision; 89.2% mAP (Private) | Effective feature focus; sensitive to occlusion and lighting |
| Ai et al. [69] | DACNN (Dual Attention CNN: Spatial + Channel Attention) | 95.4% accuracy (Private) | Strong subtle behavior discrimination; architectural complexity limits edge use |
| Li et al. [70] | CNN + SWAM (Spatial Weighted Attention Module) | 93.97% accuracy (AUC V2) | Long-range visual focus; improved interpretability |

*Summary and Comparative Analysis of Visual Approaches*

Visual-based distracted driving detection has evolved through a variety of architectural innovations, including CNNs, temporal models, ensemble/multi-task frameworks, attention-enhanced networks, and, to a lesser extent, traditional machine learning. CNNs remain foundational, with region-specific, lightweight, and transfer learning variants achieving accuracy ranging from 86.1% to 99.93% (Figure 3, Table 1). However, CNNs in this context typically operate on single frames, limiting their ability to detect distractions that unfold temporally or cognitively.

Temporal models such as 3D CNNs, CNN-RNN hybrids, and Transformers address this limitation by modeling sequential dependencies, achieving performance between 96% and 99.8% under controlled conditions (Figure 4, Table 2). Still, their high computational demands and reliance on annotated video sequences pose deployment challenges. Ensemble and multi-task models improve robustness and generalizability through architectural diversity and shared learning objectives, with accuracy ranging from 92% to 99.81% (Figure 5, Table 3). Yet, they remain sensitive to overfitting and require substantial computational resources.

Attention-based models further refine feature learning by emphasizing salient regions like the face and hands, improving accuracy and interpretability (91.6%–95.6%) (Figure 6, Table 4). However, architectural complexity and insufficient validation under varied environmental conditions limit their practical adoption. Traditional ML methods such as ADW-KNN offer lightweight alternatives with decent performance (94.04%) but lack scalability and flexibility across real-world contexts [71].

Overall, while visual models demonstrate high accuracy in controlled datasets, their generalization suffers in unconstrained environments. Most approaches rely heavily on appearance-based features, neglecting cognitive, contextual, or multimodal cues critical for comprehensive distraction detection. These findings highlight the need for integrating richer data streams to address these limitations.

**Limitations of Visual-Only Approaches and the Need for Complementary Modalities**

Despite the clear strengths of visual data approaches, including high classification accuracy, spatial feature richness, and rapid detection capabilities, these methods are inherently limited by their reliance on visual appearance alone. They often fail to account for cognitive distractions or contextual cues not captured by driver-facing cameras, such as sudden braking, lane drift, or internal vehicle dynamics. Moreover, generalizability remains a concern, particularly across diverse lighting conditions, driver demographics, and driving environments [72], [73]. These limitations underscore the need for additional sensing modalities that can complement visual data with behavioral, physiological, and vehicular context. The next section explores sensor-based approaches, which utilize inputs such as inertial signals, steering behavior, and vehicle telemetry to provide a more holistic representation of distraction in real-world scenarios.

*B. Sensor (Telemetry) Data Approaches*

Visual data have dominated distracted driving detection research, and their limitations particularly in capturing internal cognitive states or subtle physical cues highlight the need for complementary data sources. Sensor-based or telemetry-driven approaches provide an alternative by leveraging real-time signals from vehicle-embedded systems and wearable technologies. These include Inertial Measurement Units (IMUs), GPS, accelerometers, gyroscopes, steering angle sensors, and even physiological signals like EEG. Unlike visual data, which focus primarily on spatial features captured from driver-facing cameras, sensor data capture dynamic behavioral patterns, vehicular control inputs, and environmental context over time.

This section reviews methods that use sensory inputs to model driver distraction, spanning both machine learning (ML) and deep learning (DL) techniques. The DL segment includes RNNs, LSTM models, CNN-RNN hybrids, and attention-enhanced architectures, while the ML portion focuses on traditional classifiers built on engineered sensor features. These models aim to detect driver distraction more holistically extending beyond what can be inferred from visual appearance alone—by capturing temporal dynamics, control patterns, and physiological responses.

Figure 7 gives an overview of the various sensory data approaches captured in this review.

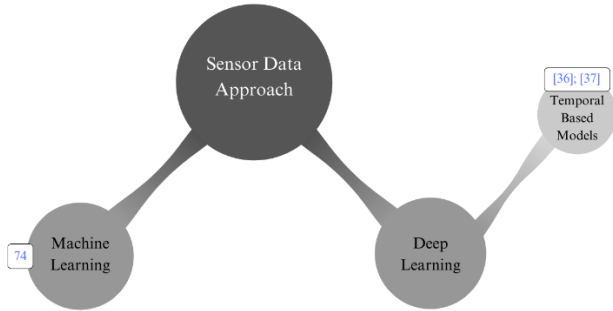

**Fig. 7.** Summary of Sensory Data Approaches

*ML and DL-Based Approaches*

**Machine Learning Approaches**

ML methodologies have been implemented for the analysis of sensor data in distracted driving detection, providing computationally efficient and interpretable alternatives to deep learning architectures. Richter et al. [74] introduced the Phone-Locating-Unit (PLU), which utilized in-cabin cellular signal patterns in conjunction with inertial measurement data processed through neural network frameworks to differentiate between driver and passenger phone usage. Their system demonstrated an impressive 98.4% classification accuracy under diverse experimental conditions; however, performance efficacy decreased significantly when attempting to distinguish driver-right-hand phone usage in proximity to the gear shift mechanism. These performance limitations underscore the persistent challenges associated with signal interference patterns and contextual ambiguity in real-world driving environments. While ML approaches exhibit considerable promise in resource-constrained computational environments, their generalizability across diverse driving scenarios remains fundamentally limited by the inherent constraints of manual feature engineering processes and variability in sensor characteristics and placement.

**Deep Learning Approaches**
**Temporal Models**

Temporal models play a crucial role in sensor-based distracted driving detection by capturing sequential dependencies across time-series data streams such as steering angle, vehicle velocity, and physiological signals. Unlike static classifiers, these architectures can recognize distraction patterns that evolve over time, providing improved contextual awareness and prediction robustness.

Kouchak and Gaffar [36] implemented two RNN-based architectures—Bidirectional LSTM (Bi-LSTM) and LSTM with an attention layer—on simulated driving datasets capturing lane position, velocity, and steering inputs. The attention-enhanced LSTM outperformed the Bi-LSTM in both training and testing, achieving lower mean absolute error (MAE: 0.85 training, 0.96 testing) compared to the Bi-LSTM (0.97 training, 1.00 testing). The attention layer enabled the model to prioritize the most relevant sequence segments, improving interpretability and computational efficiency. However, the reliance on simulated environments raised concerns about ecological validity, and the input space was limited to vehicular control metrics, excluding richer contextual or cognitive indicators. Furthermore, scalability to embedded platforms was not explored.

Yan et al. [37] advanced the temporal modeling landscape by integrating EEGNet for spatial feature extraction from electroencephalography (EEG) signals with an LSTM module to model time-dependent neural activity. Conducted in a 2-back paradigm simulating visual and auditory distractions, the model achieved 71.1% accuracy across three classes (focused, visual distraction, auditory distraction). Notably, performance remained stable when reducing the number of EEG channels from 63 to 14, demonstrating computational efficiency with minimal accuracy loss. Nonetheless, the simplified driving context and reliance on EEG alone limited generalization to more complex, multimodal real-world settings. The authors advocated for a multibranch architecture to better disentangle overlapping neural patterns across distraction types.

Together, these studies demonstrate the utility of temporal models in sensor-based distraction detection, particularly when combining recurrent structures with attention mechanisms or spatial encoders. While results highlight promising accuracy and efficiency trade-offs, broader validation on real-world driving datasets and fusion with complementary modalities remain key avenues for future research.

Figure 8 provides a generalized pipeline of sensor-based approaches, and Table 5 summarizes performance.

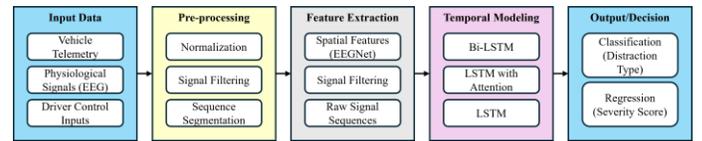

**Fig. 8.** Generalized pipeline for temporal modeling using sensor (telemetry) data in distracted driving detection.

TABLE 5
SUMMARY OF TEMPORAL BASED MODEL APPROACHES

| Study | Architecture / Model Type | Performance | Key Notes |
|---|---|---|---|
| Kouchak and Ghaffar [36] | LSTM with Attention | MAE 0.85 (train), 0.96 (test) | Improved temporal focus; attention reduced input noise and boosted real-time applicability |
| Yan et al. [37] | EEGNet + LSTM | 71.1% accuracy (3-class) | Helped distinguish learning effects with lower EEG channel use |

*Summary and Comparative Analysis of Sensory Approaches*

Sensor-based approaches offer a complementary lens to visual models by capturing internal vehicle dynamics, control inputs, and physiological signals that reflect driver behavior over time. Across both ML and DL methods, these approaches have demonstrated considerable potential in detecting distraction types that may not manifest visually—such as cognitive or motor distractions. Richter et al. [74], excel in low-latency, real-time applications but face scalability issues due to reliance on hand-crafted features and limited contextual depth.

Deep learning models, particularly temporal architectures like LSTM and Bi-LSTM, have proven effective in modeling sequential driving behavior using telemetry inputs such as steering angle, lane deviation, and velocity. Attention mechanisms further enhance performance by dynamically weighting signal importance, as seen in Kouchak and Gaffar's [36] LSTM-with-attention model, which reduced mean absolute error compared to traditional recurrent structures. Moreover, hybrid frameworks such as Yan et al. [37]'s EEGNet-LSTM model bridge spatial and temporal modeling, achieving 71.1% accuracy in classifying visual and auditory distractions from EEG signals. While this illustrates the utility of physiological telemetry, performance lags visual-only models, and real-world generalizability remains limited due to reliance on simulated driving data.

Accuracy in sensor-based models generally ranges from 71% to 98.4%, with real-time feasibility improving as models integrate fewer channels or lightweight architectures. However, common limitations include narrow data modalities (e.g., steering and speed alone), constrained input diversity, and insufficient robustness under real-world driving conditions. Compared to visual systems, sensor-based approaches offer improved coverage of cognitive and internal distraction types but often lack the spatial context needed to differentiate fine-grained driver actions.

These observations underscore the complementary nature of sensor and visual modalities. While neither modality alone offers a complete representation of driver distraction, their integration in multimodal systems presents a promising path forward. By fusing spatial visual cues with temporal telemetry patterns and physiological signals, future architectures can achieve more robust, generalizable, and context-aware distraction detection. The next section explores such multimodal frameworks and the fusion strategies that underpin them.

*C. Multimodal Approaches*

Distracted driving is inherently multifaceted, influenced by not only visual behavior but also cognitive load, physiological state, and vehicular control dynamics. While unimodal systems, such as in-cabin video or vehicle telemetry, have demonstrated high classification accuracy under controlled conditions, they often lack contextual grounding. For instance, a stationary car in a parking garage with the driver adjusting the radio may be flagged as "distracted" without additional sensory input. This limitation can lead to false positives or missed detections, especially in ambiguous cases involving subtle postural shifts or brief glances.

Multimodal approaches aim to overcome these challenges by integrating heterogeneous data streams, including driver-facing video, inertial measurements, physiological signals (e.g., EEG, heart rate), and vehicle dynamics (e.g., speed, steering angle). By cross-validating cues across modalities, these systems improve robustness, enhance interpretability, and offer greater generalization across real-world driving scenarios.

Fusion strategies are commonly categorized into early fusion (combining raw or low-level features), late fusion (aggregating independent model outputs), and hybrid fusion (capturing both shared and modality-specific representations) [14]. Deep learning models such as CNN-RNN hybrids, multi-branch attention modules, and transformers have gained traction in this domain due to their capacity to model complex temporal and cross-modal dependencies. Figure 9 presents the techniques applied to multimodal data.

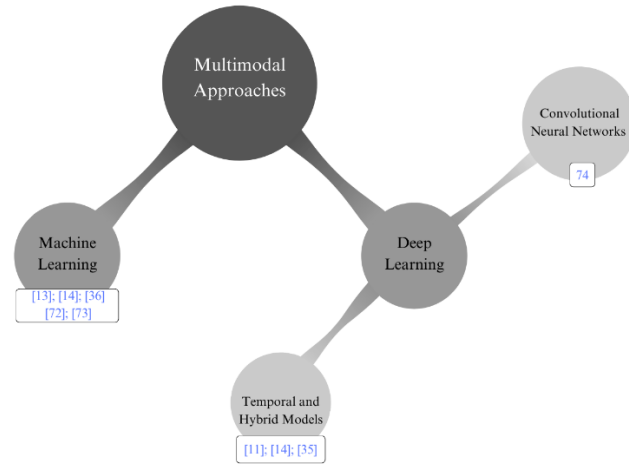

**Fig. 9.** Representative Multimodal Techniques

*ML and DL-Based Approaches*

**Machine Learning Approaches**

ML approaches for multimodal distracted driving detection typically integrate handcrafted features from visual, physiological, and vehicle-based modalities. Studies often converge around combining classical classifiers like Random Forest (RF), Support Vector Machines (SVM), and Extreme Gradient Boosting (XGB) with data streams such as heart rate variability, facial movement, and vehicle telemetry.

A common trend involves fusing visual and physiological features for improved accuracy. Gjoreski et al. [14] and Das et al. [13] both achieved high F1-scores (94%) by combining facial action units, emotions, and thermal imaging with electrodermal activity or blood volume pulse features. Their results reinforce the idea that physiological cues can amplify detection precision, especially for physical and affective distractions. However, both studies revealed scalability issues due to class imbalance and reduced sensitivity to cognitive distractions.

In contrast, Misra et al. [39] and Heenetimulla et al. [75] focused more on integrating eye tracking and ECG signals with simpler classifiers like SVMs or FPGA-optimized pipelines. While Misra's model reached 90% accuracy, the system was limited by short data collection periods and lack of individualized modeling. Heenetimulla's real-time implementation achieved 93.49% accuracy for phone use but struggled with driver health monitoring (78.69 percent), revealing a trade-off between specificity and generalizability when relying on wearable sensors.

Vehicle telemetry features also proved valuable in Yadawadkar et al. [76], who leveraged head pose and steering data from the SHRP2 dataset. Their study highlighted the importance of synthetic oversampling (SMOTE) to balance datasets and achieved a 94% F1-score. Still, limited training data and short temporal sequences hindered model robustness, prompting suggestions like meta-learning and PerClos-based eye metrics for improvement.

Overall, ML-based multimodal systems offer practical solutions for real-time distraction detection using lightweight classifiers and interpretable features. Yet, they face challenges with generalization, data diversity, and sensor dependence. These limitations underscore the importance of scalable data fusion strategies and richer contextual modeling to enhance performance beyond controlled environments.

Figure 10 shows fusion strategies in ML-driven multimodal distraction detection typically fall into three categories: early fusion, where features from all modalities are concatenated before modeling; mid fusion, which integrates modality-specific representations at intermediate network layers; and structured fusion (e.g., STRNet), which employs residual and modality-specialized blocks for hierarchical integration. These strategies reflect the architectural diversity used to combine visual, physiological, and vehicular data streams in machine learning studies. The STRNet model by Gjoreski et al. [14], for example, exemplifies structured mid-level fusion tailored to preserve both shared and modality-specific information. Table 6 presents key summary statistics for machine learning based approaches to multimodal data.

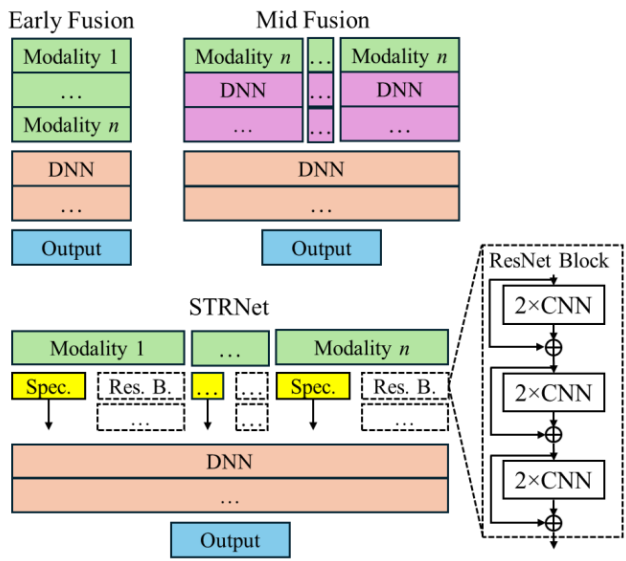

**Fig. 10.** Representative Fusion Architectures in Multimodal Machine Learning for Distracted Driving Detection Adapted from Gjoreski et al. [14].

TABLE 6.
SUMMARY OF MACHINE LEARNGIN MULTIMODAL MODEL APPROACHES

| Study | Modality Used | Method | Performance | Key Notes |
|---|---|---|---|---|
| Gjoreski et al. [14] | Visual (Emotions, AUs), Physiological (EDA, BVP) | XGBoost Gradient Boosting | F1-score: 94% (Private Dataset) | Supervised fusion; tuned for nuanced distraction |
| Das et al. [13] | Thermal, Video, Physiological (EDA, BVP, GSR) | Random Forest, SVM | Accuracy: 94% (Simulated Driving) | Affected by participant diversity, class imbalance |
| Heenetimulla et al. [75] | Eye Tracking, ECG, Thermal Video | Ensemble Tree Classifiers | 93.49% (Phone); 78.69% (Health Monitoring) | Real-time; struggled with wearable-based health distraction detection |
| Misra et al. [39] | Eye Tracking, ECG, Kinematics | Random Forest, SVM | 90% (Private Dataset) | Short data window; lacked personalization |
| Yadawadkar et al. [76] | Visual (Head Pose), Vehicle (Speed, Lane, Steering) | ML/DL Feature Fusion + ML Classifiers | F1-score: 94% (SHRP2NDS Subset) | SMOTE balancing; task-specific contextual features |

### Deep Learning Approaches
#### Convolutional Neural Networks

Martin et al. [77] introduced the Drive&Act dataset, a large-scale multimodal benchmark that includes synchronized RGB, infrared, depth, and 3D skeleton data across 83 annotated driver behaviors. Using CNN-based models such as C3D, P3D ResNet, and I3D, they demonstrated that I3D achieved the highest accuracy (63.64%), which improved to 69.03% with late fusion across modalities and views. These findings underscore the strength of integrating visual and sensor data in CNN frameworks to overcome challenges such as poor illumination, occlusions, and subtle driver behavior cues. However, the study was conducted in a simulator, raising concerns about ecological validity and real-world scalability. Additionally, the reliance on hardware like Kinect v2 limits deployment flexibility.

Figure 11 illustrates the generalized architecture of Martin et al.'s multimodal CNN framework, showing the integration of spatiotemporal features and late fusion for final behavior classification.

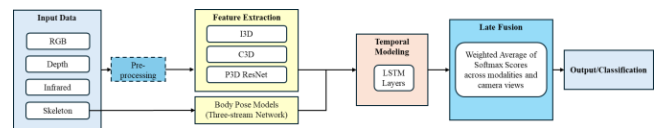

**Fig. 11.** Generalized pipeline of the multimodal CNN-based framework. Adapted from Martin et al. [77] study.

#### Temporal Models

Temporal deep learning models have proven effective in modeling time-evolving distractions that span beyond frame-level perception. Gjoreski et al. [14] employed the Spectro-Temporal ResNet (STRNet), an end-to-end architecture that fused visual features with physiological signals, achieving an F1-score of 87%. STRNet's design was particularly adept at capturing temporal dependencies across heterogeneous modalities with differing signal frequencies (e.g., facial motion and heart rate). This work emphasized the advantages of combining spectral representations with temporal convolution, as depicted in Figure 10, which illustrates STRNet's multi-branch residual and modality-specific processing blocks.

#### CNN-RNN Hybrids

Hybrid models that combine convolutional and recurrent layers have shown promising results in capturing both spatial and temporal characteristics of multimodal data. Omerustaoglu et al. [11] proposed a CNN-LSTM hybrid framework where visual features extracted via VGG16 and Inception V3 were combined with vehicle telemetry data such as engine RPM and throttle position. Their experiments with both feature-level and decision-level fusion strategies led to a 23% improvement in classification accuracy, peaking at 85%. However, the dataset consisted of data from a single driver, limiting the system's demographic generalizability and real-world application potential.

In pursuit of computational efficiency, Anagnostou and Mitianoudis [38] developed a lightweight ConvGRU model that fused RGB, infrared, and depth data for real-time distraction detection. Despite operating at just 24.4% of the computational cost of 3D ResNet18, the model achieved an F1-score of 0.7205 and an AUC of 0.8866 on the DAD dataset. Nonetheless, its performance lagged heavier models, particularly under class imbalance conditions. The absence of ensemble strategies and the use of relatively homogeneous datasets further restricted its generalizability.

Figure 12 presents a simplified CNN-RNN hybrid architecture showing the fusion of visual and telemetry inputs for behavior classification.

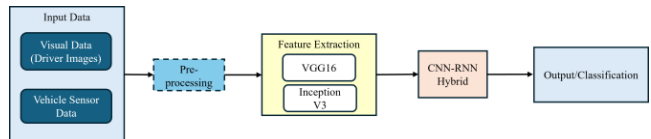

**Fig. 12.** A simplified CNN-RNN hybrid architecture integrating visual and vehicle sensor data for distraction classification.

*Summary and Comparative Analysis of Multimodal Approaches*

Multimodal approaches have demonstrated substantial promise in distracted driving detection by integrating visual, physiological, and telemetry data streams. These systems address the limitations of unimodal models—particularly their reliance on appearance-based cues or isolated sensor signals—by fusing heterogeneous modalities to enhance robustness, contextual awareness, and generalization to real-world driving scenarios.

Several studies benchmarked their multimodal architectures against unimodal baselines and reported consistent performance gains. For instance, Omerustaoglu et al. [11] showed that integrating CNN-extracted visual features with telemetry inputs via LSTM improved classification accuracy by 23%, underscoring the value of data fusion in capturing complex driver behaviors. Similarly, Martin et al. [67] reported that late fusion of RGB, infrared, depth, and skeleton modalities using an I3D backbone increased accuracy from 63.64% to 69.03%, illustrating how modality complementation can mitigate challenges like occlusion and poor illumination.

Beyond performance improvements, multimodal frameworks broaden the scope of distraction detection. Das et al. [13] achieved a 94% F1-score in recognizing physical and frustration-based distractions by combining thermal, visual, and physiological features—categories where unimodal models often struggle due to class imbalance or weak signal discrimination. Misra et al. [39] achieved 90% accuracy in detecting cognitive distractions by fusing heart rate variability, eye-tracking, and vehicle dynamics, a feat rarely attained through visual data alone.

Moreover, Gjoreski et al. [14] demonstrated that their Spectro-Temporal ResNet (STRNet), trained on visual and physiological signals, outperformed traditional machine learning methods applied to single modalities, achieving an F1-score of 87%. This result underscores the advantage of deep learning-driven multimodal fusion for end-to-end behavior recognition.

Despite these gains, multimodal systems present new challenges, including increased computational demands, synchronization overhead, and reliance on high-fidelity or wearable sensors. For example, Heenetimulla et al. [75] highlighted the dependence on smartwatches and the reduced accuracy of heart rate variability monitoring (78.69%) in driver health assessment, limiting real-world scalability. Similarly, the limited participant pool in Omerustaoglu et al. [11] raises concerns about demographic generalizability.

Nonetheless, empirical evidence consistently shows that multimodal systems outperform unimodal counterparts in benchmarked scenarios—particularly for nuanced distraction detection and real-time driver state estimation. Their performance advantage, combined with enhanced contextual reasoning, positions multimodal architectures as a pivotal direction for robust, scalable, and interpretable distracted driving detection systems.

*D. Emerging Approaches*

As research in distracted driving detection advances, new modalities and architectural innovations continue to emerge that challenge the dominance of visual and telemetry-based methods. These approaches often aim to address key limitations such as privacy, intrusive sensing, and limited contextual awareness. This section highlights novel detection paradigms that leverage neuro-inspired computation, radio frequency sensing, and auditory signal processing to extend the frontier of distraction detection capabilities.

*Neuro-Inspired Models*

Shariff et al. [78] introduced a spiking neural network (SNN) architecture for distraction detection using event-based data from the simulated DMD (v2e) dataset. By mimicking biological neuron firing patterns, the Spiking-DD model handled high temporal resolution data with low parameter complexity, offering real-time readiness and computational efficiency. A privacy-by-design framework further enhanced its applicability in contexts where visual monitoring may raise ethical concerns. The model outperformed traditional architectures in both accuracy and resource demands, demonstrating the promise of neuromorphic computing for driver monitoring. Nonetheless, its reliance on simulated environments limits external validity. The model's performance under real-world lighting and motion variance remains untested, and its deployment potential on neuromorphic hardware like Intel's Loihi-2 has yet to be fully realized. These gaps suggest a critical need for in-the-wild validation and hardware benchmarking before broader adoption. Figure 13 represents the novel neuro-inspired approach by Shariff et al. [78].

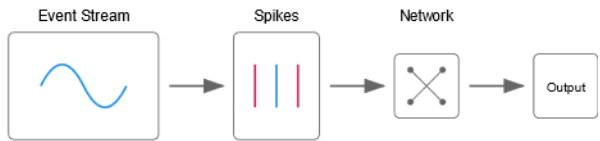

**Fig. 13.** Overview of sensing pipeline adapted from Shariff et al. [78].

*Radio Frequency – Based Methods*

Raja et al. [79] proposed a non-intrusive, Wi-Fi-based distraction detection system utilizing Channel State Information (CSI) to capture upper body motion, including head turns and arm gestures. Operating through a single transmitter-receiver pair, the system analyzed phase, and subcarrier features of Wi-Fi signals to classify distraction behaviors with 94.5% accuracy. As a vision-free alternative, this radio frequency approach presents strong potential for privacy-preserving, scalable deployment in everyday vehicles.

However, the use of modified WLAN cards and the lack of evaluation in multi-occupant environments limit the method's practicality and scalability. Future research must integrate manufacturer support for CSI data access and extend validation to more complex cabin scenarios. Figure 14 shows the general flow for the proposed model presented by Raja et al. [79].

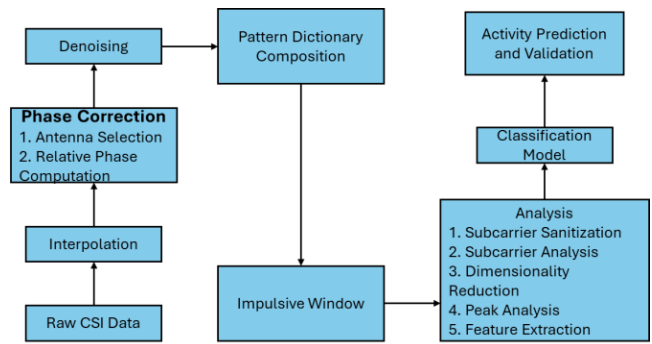

**Fig. 14.** Proposed System Architecture for Wi-Fi-based distraction. Adapted from Raja et al. [79].

*Auditory Signal Processing*

Zhao et al. [80] developed a novel auditory-based distraction detection system that leverages wearable acoustic sensors to capture under-skin neck vibrations. By converting these signals into Mel-spectrograms and feeding them into a ResNet50V2 classifier, the system identified behaviors such as coughing, sneezing, talking, and yawning with an F1 score of 91.32%. The approach offers a visually unobtrusive, privacy-friendly alternative for distraction monitoring, particularly useful in scenarios where video or touch-based sensing may be limited or intrusive.

Despite its strengths, the model was trained on a narrow set of distraction types and lacks validation on larger, more diverse datasets. Its applicability for broader behavioral or health monitoring remains an open avenue for future development. Zhao et al.'s architecture is presented in Figure 15.

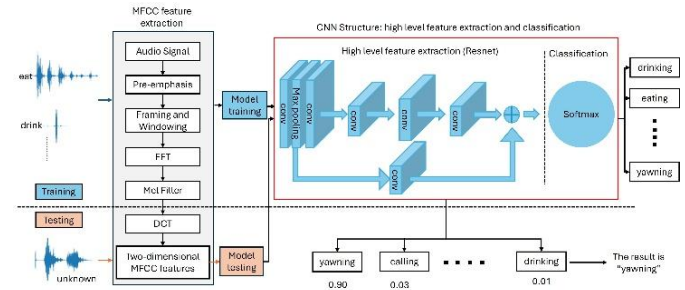

**Fig. 15.** Overall architecture of the proposed methodology adapted from Zhao et al. [80].

## IV. CONCLUSION

This systematic review identifies a critical imbalance in modality focus within distracted driving detection research and reveals key opportunities for advancing real-world, context-aware systems through modality fusion, robust validation, and practical deployment frameworks. While significant strides have been made in algorithmic sophistication and classification accuracy, our analysis underscores persistent challenges in building robust systems that generalize beyond benchmark conditions and support real-time deployment.

*Key Findings*

**Visual Modality Dominance and its limitations**

The review clearly establishes the *predominance of visual modality approaches* in distracted driving detection research [3], [20], [21]. Visual systems have demonstrated impressive accuracy metrics (86.1–99.93%) through architectures such as transfer-learned CNNs [37], [39], attention-enhanced models [50], [58], [60], and temporal frameworks [34], [51], [52]. However, despite their high reported performance on benchmark datasets, visual-only systems remain fundamentally limited in four critical areas:

- Limited robustness to lightning, occlusion, and environmental variability [29], [48], [50].
- Inability to detect cognitive distractions not reflected in physical appearance [42], [40], [14].
- Lack of contextual grounding without sensor or vehicle integration [11], [13].
- High computational demands that hinder edge deployment feasibility [31], [52], [35], [33].

These limitations highlight a disconnect between laboratory metrics and real-world performance, emphasizing the need for evaluation frameworks that move beyond accuracy alone.

**Alternative Modalities: Emerging Promise**

While less explored, alternative modalities provide critical complementary capabilities:

- Sensor-based models capture temporal dynamics and cognitive load with up to 98.4% accuracy [61], [62], [73], but often rely on simulated environments and manual feature engineering.
- Auditory-based systems offer unobtrusive, privacy-friendly distraction detection with F1-scores exceeding 91% [74].
- Multimodal systems enhance robustness and generalization, but unimodal visual models still achieve top performance under controlled conditions [13], [14], [67], [11], [65], [81].

**Methodological and Evaluation Gaps.**

The review also identifies several critical research gaps that hinder real-world adoption:

- Insufficient cross-dataset testing reduces confidence in model generalizability [52], [57], [67].
- Overreliance on simulation limits ecological validity [62], [65], [67].
- Inconsistent evaluation metrics obscure true performance trade-offs [56], [68].
- Limited attention to privacy and ethical design across most implementations [14], [63], [69].

**Future Research Directions**

*Modality Integration and Fusion*

- Increasing research focuses on underutilized modalities and cross-modal integration [14], [67].
- Design adaptive fusion frameworks that weight inputs based on distraction context [58], [60].
- Leverage self-supervised learning to reduce data annotation bottlenecks and enhance generalization [21].

Context-Aware Detection Frameworks

- Incorporate external variables (road, weather, traffic) to enrich model context [66], [13].
- Establish personalized driver baselines to improve sensitivity and reduce false positives [64], [65].
- Model distraction patterns over extended temporal sequences to detect evolving behaviors [51], [73].

Practical Deployment Considerations

- Pursue lightweight models for real-time inference on embedded systems [33], [35], [59].
- Adopt privacy-preserving frameworks such as federated learning or neuromorphic sensing [63], [69], [74].
- Integrate explainability into AI outputs to support driver feedback and regulatory compliance [82], [60].

Evaluation and Validation Frameworks

- Develop cross-modality datasets with diverse driving contexts and user profiles [67], [40].
- Establish rigorous real-world testing protocols under variable conditions [62], [67].
- Adopt multidimensional evaluation frameworks measuring latency, energy, and distraction severity [54], [57], [83].

*Summary and Outlook*

This review demonstrates that while significant advances have been made in distracted driving detection through machine learning approaches, substantial challenges remain in developing truly robust, generalizable systems suitable for real-world deployment. The overemphasis on visual-only approaches represents a critical limitation, especially given the growing evidence of the advantages offered by sensor-based, auditory, and multimodal frameworks [13], [14], [67].

Future research must prioritize holistic, multimodal integration, deployable context-aware architectures, and rigorous real-world validation to bridge the gap between benchmark performance and on-road reliability. These advancements will not only improve driver monitoring technologies but also provide the foundation for scalable, ethical, and impactful road safety interventions.

*These insights can inform not only technological innovation but also regulatory guidelines for in-vehicle monitoring systems and driver safety legislation [3], [15].* By aligning detection technologies with practical constraints and ethical considerations, next-generation systems can meaningfully reduce distraction-related crashes and contribute to safer roads globally.

ACKNOWLEDGMENT

The authors would like to thank the Southeast Transportation Workforce (SETWC) for their support.

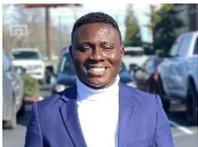
**Anthony Dontoh** is a doctoral candidate in Civil Engineering at the University of Memphis and serves as a graduate research assistant at the Southeast Transportation Workforce and a teaching assistant at the Department. His current research focuses on intelligent transportation systems, road safety, distracted driving detection and multimodal analysis. He has also contributed to a high level impact project in advancing gender equality in the transportation sector.

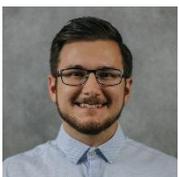
**Logan Sirbaugh**, EI, is a doctoral candidate in Civil Engineering at the University of Memphis and serves as an Instructor in the Herff College of Engineering. He received his B.S. and M.S. degrees in Civil Engineering from the University of Memphis. His research interests include active transportation systems, road ecology, and transportation workforce development, with particular emphasis on bicycle level of traffic stress frameworks and naturalistic driving studies. He is currently a Graduate Research Assistant with the Southeast Transportation Workforce Center, where he has contributed to multiple projects for federal, state, and local transportation agencies.

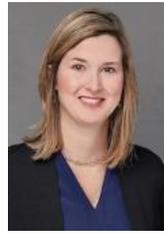
**Dr. Stephanie Ivey** is a Professor of Civil Engineering and Director of the Southeast Transportation Workforce Center in the Herff College of Engineering at the University of Memphis. Her research focuses on transportation planning, operations, and workforce development. She is a member of the Federal Reserve Bank of St. Louis Transportation Industry Council and co-chair of the Institute of Transportation Engineers STEM Committee.

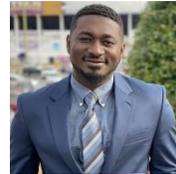
**Dr. Armstrong Aboah** (Member, IEEE) received his B.S. in Civil Engineering from Kwame Nkrumah University of Science and Technology (2017), M.S. from Tennessee Technological University (2019), and Ph.D. from the University of Missouri (2022), specializing in transportation engineering with computer vision and machine learning. His research on intelligent transportation systems spans computer vision, transportation sensing, and deep learning, with over 15 peer-reviewed publications in venues including CVPR and NeurIPS. Dr. Aboah has secured competitive funding and pioneered work in vision-based traffic anomaly detection, helmet violation detection, and pavement roughness estimation.

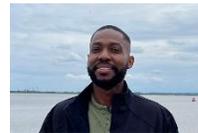
**Andrews Danyo** is a Civil Engineer with a passion for innovation and smart initiatives. He also has a background in Construction Management, Artificial Intelligence and Data Science. His work focuses on creating smarter transportation systems and urban infrastructure that serves people, not just statistics. What drives Andrew is seeing how intelligent systems can solve real-world infrastructure challenges or develop sustainable solutions that adapt to human needs. He brings practical knowledge of how things get built and the vision for how technology can make them work better for everyone.